\documentclass{IEEEtran}
\usepackage{cite}
\usepackage{amsmath,amssymb,amsfonts}
\usepackage{graphicx}

\usepackage{textcomp,nicefrac}
\usepackage{multirow}
\usepackage{array}
\usepackage{url}
\usepackage{flafter}
\usepackage{placeins}
\usepackage{capt-of}
\usepackage[T1]{fontenc}
\def\BibTeX{{\rm B\kern-.05em{\sc i\kern-.025em b}\kern-.08em T\kern-.1667em\lower.7ex\hbox{E}\kern-.125emX}}

\begin{document}

\raggedbottom
\title{Joint Initialization of Flux Networks and Effective Multiplication Factor for Physics-Informed Neural Networks Solving Neutron Diffusion Problems}
\author{Qin Hang, Yangdi Yi, Jiayi Li, Xu Wang, and Heng Zhang%
\thanks{This work was supported by the Science and Technology on Reactor System Design Technology Laboratory (Grant No. LRSDT12023108) and the Research Start-up Fund Project of Chongqing University of Posts and Telecommunications under Grant Nos. A2020-217 and A2020-216.}%
\thanks{Qin Hang, Yangdi Yi, Jiayi Li, Xu Wang, and Heng Zhang are with the School of Computer Science and Technology, Chongqing University of Posts and Telecommunications, Chongqing 400065, China.}%
\thanks{Qin Hang, Jiayi Li, and Heng Zhang are also with the Center for Scientific Intelligence Innovation, University of Science and Technology of China, Hefei 230026, China.}%
\thanks{Qin Hang and Heng Zhang are also with the Institute of Advanced Technology, University of Science and Technology of China, Hefei 230088, China.}%
\thanks{Corresponding author: Heng Zhang (e-mail: zhangheng@cqupt.edu.cn).}}
\maketitle

\begin{abstract}
Efficient determination of the effective multiplication factor (\(k_{eff}\)) is an important computational task in reactor core neutronics analysis. Physics-informed neural networks (PINNs) incorporate neutron diffusion equations and boundary conditions into network training to efficiently determine the neutron flux distribution and \(k_{eff}\). To further improve the efficiency of \(k_{eff}\) calculations using PINNs, a Joint Initialization Physics-Informed Neural Network (JI-PINN) is proposed in this work. In this method, a low-resolution approximate solution to the K-eigenvalue problem is used to construct a joint initial state for the flux network parameters and \(k_{eff}\), and both are then jointly optimized under physical constraints. The proposed method was validated on a two-dimensional two-group two-material case, the IAEA 2D benchmark, a two-dimensional two-group four-material case, and a three-dimensional single-group case. For these test cases, the total computational time was reduced by 25.4\%, 38.2\%, 49.4\%, and 28.9\%, respectively, while comparable solution accuracy was maintained. The occurrence of anomalous results associated with marked deviations of \(k_{eff}\) from the reference value was also reduced. The proposed method provides a more efficient and robust initialization strategy for solving neutron diffusion K-eigenvalue problem with PINNs.
\end{abstract}


\section{Introduction}
\label{sec:introduction}
Efficient determination of the effective multiplication factor (\(k_{eff}\)) is an important task in reactor core neutronics analysis. As a key parameter in reactor criticality analysis, the calculated value of \(k_{eff}\) is used to determine whether a reactor system is subcritical, critical, or supercritical \cite{ref1,ref2,ref3}. For a given core geometry, material parameters, and boundary conditions, \(k_{eff}\) and the corresponding neutron flux distribution need to be determined. The resulting \(k_{eff}\) and neutron flux distribution provide an important basis for reactor criticality analysis and core power distribution calculations.

In current practice, spatial discretization methods such as the finite difference method, finite element method, and nodal method are commonly combined with power iteration to determine the neutron flux distribution and \(k_{eff}\). These methods are well established and achieve reliable numerical accuracy \cite{ref4,ref5}. However, traditional numerical methods generally require spatial discretization of the computational domain and repeated iterations to update the neutron flux and \(k_{eff}\). Their computational cost may therefore remain high for high-dimensional problems. In recent years, with the development of deep learning, neural network approaches have been increasingly applied to reactor analysis \cite{ref6,ref7}. Meanwhile, their use for approximating solutions to partial differential equations (PDEs) has attracted increasing attention. Among these approaches, PINNs incorporate governing equations and boundary conditions into neural network training, and PDE solutions are approximated through optimization of the neural network parameters \cite{ref8,ref9}.

Building on these developments, PINNs have been applied to solve neutron diffusion equations. Zhang et al. proposed GHO Hybrid PINN and improved the performance in solving multidimensional, multigroup neutron diffusion problems through modifications to the network architecture and optimization strategy \cite{ref10}. Wang et al. developed a surrogate model for neutron diffusion and demonstrated the feasibility of PINNs for solving neutron diffusion problems \cite{ref11}. Do et al. investigated a PINN method based on a mixed dual formulation for solving neutron diffusion equations, further extending the application of PINNs to different equation formulations \cite{ref12}. Zhang et al. proposed $R^{2}$-PINN by combining a convolutional neural network with residual adaptive resampling for solving neutron diffusion equations \cite{ref13}.

For \(k_{eff}\) determination in neutron diffusion equations, Elhareef and Wu extended PINNs to K-eigenvalue problems by treating \(k_{eff}\) as a trainable parameter and imposing a regularization constraint to exclude the trivial zero-flux solution \cite{ref14}. Y. Yang et al. proposed a data-enabled physics-informed neural network (DEPINN) and introduced a small amount of known data into training as prior information to assist in determining the neutron flux and \(k_{eff}\) \cite{ref15}. For eigenvalue iteration and problems involving different material regions, Q.-H. Yang et al. combined the inverse power method with neural networks to propose the Generalized Inverse Power Method Neural Network (GIPMNN) and further developed the Physics-Constrained GIPMNN (PC-GIPMNN) to handle material interfaces by enforcing neutron flux continuity and neutron current continuity \cite{ref16}. Bi et al. proposed Fixed-point Constraint Physics-informed Neural Networks (FC-PINNs) to avoid trivial solutions in eigenvalue calculations and handle interface conditions between different material regions through computational domain concatenation \cite{ref17}.

However, relatively slow training remains a challenge for PINNs in solving neutron diffusion K-eigenvalue problem. One important reason lies in the random initialization of the flux network parameters and the empirical specification of the initial \(k_{eff}\). The function represented by a PINN at initialization can influence subsequent optimization behavior \cite{ref18}. Therefore, different initialization settings may lead to substantial differences in the optimization process and convergence behavior. A large discrepancy between the initial state and the target solution may increase the computational cost of model training and result in convergence difficulties or large errors in the computed \(k_{eff}\).

To address this issue, this paper proposes the Joint Initialization Physics-Informed Neural Network (JI-PINN). The proposed method uses the neutron flux distribution and the corresponding \(k_{eff}\) from a low-resolution approximate solution to the K-eigenvalue problem to construct a joint initial state for the flux network parameters and \(k_{eff}\), and then jointly updates both during subsequent physics-constrained training. The remainder of this paper is organized as follows. Section II introduces the proposed JI-PINN. Section III presents the numerical experiments and analysis of the results. Section IV summarizes this work and discusses future research directions.

\section{Methods}
\label{sec:methods}
In the steady-state neutron diffusion K-eigenvalue problem, the neutron flux in each energy group and \(k_{eff}\) are unknowns to be determined simultaneously. In this work, JI-PINN is constructed by combining a low-resolution spatial discretization approximation with PINN physics-constrained joint optimization. The overall framework is shown in Fig. 1.

\begin{figure*}[t]
\centering
\includegraphics[width=0.90\textwidth]{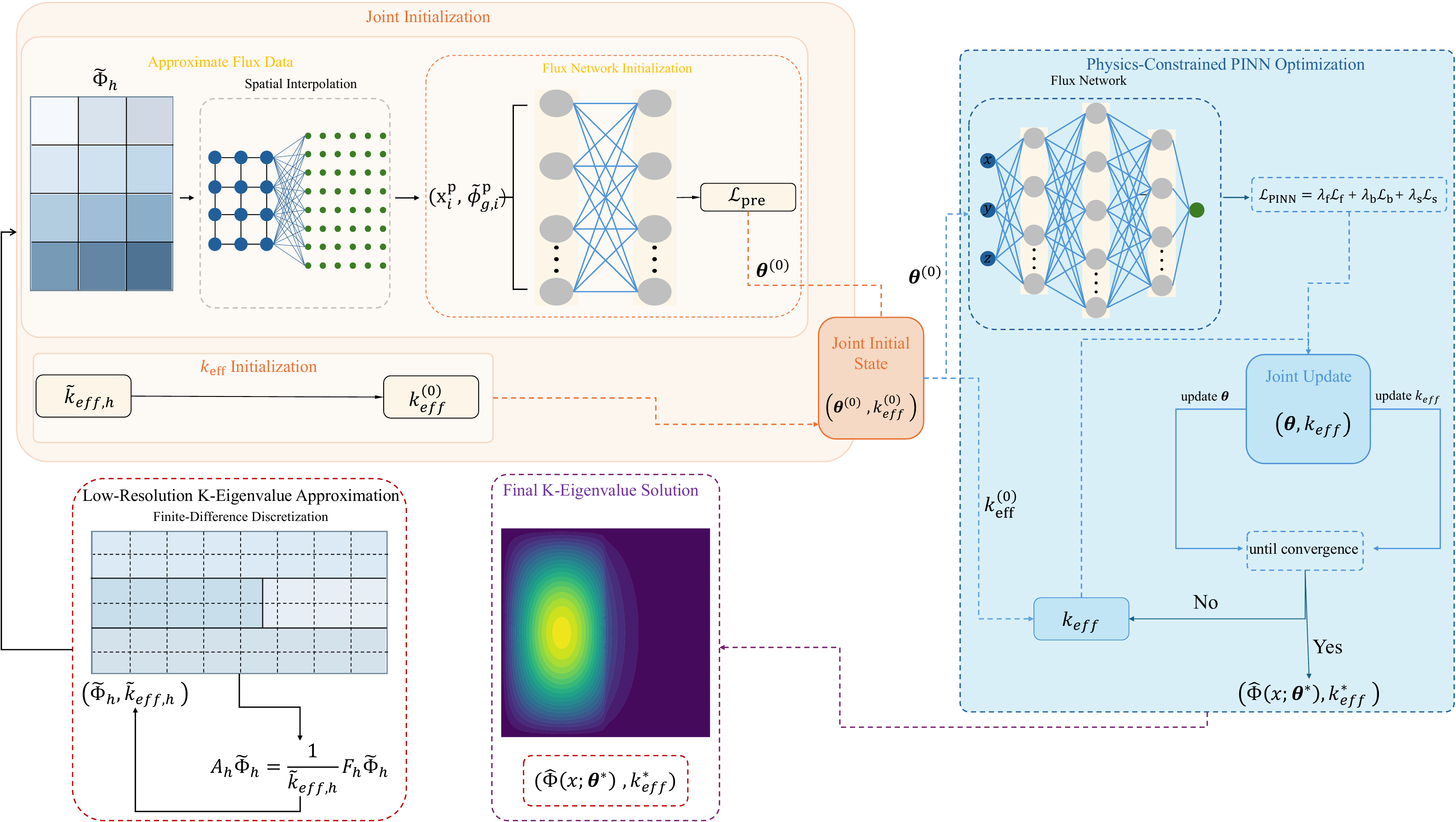}
\begingroup
\CLASSOPTIONconferencetrue
\caption{Overall framework of JI-PINN.}
\label{fig:1}
\endgroup
\end{figure*}

For a given reactor geometry, material parameters, and boundary conditions, the G-group steady-state neutron diffusion K-eigenvalue equation can be written as

\begin{equation}
\label{eq:1}
\begin{aligned}
&-\nabla \bullet \left\lbrack D_{g}\nabla\phi_{g} \right\rbrack
 + \Sigma_{\mathrm{r},g}\phi_{g}
 - \sum_{\substack{g'=1\\g'\neq g}}^{G}
   \Sigma_{\mathrm{s},g'\rightarrow g}\phi_{g'} \\
&\qquad = \frac{\mathcal{X}_{g}}{k_{eff}}
 \sum_{g'=1}^{G}\nu\Sigma_{\mathrm{f},g'}\phi_{g'},
 \quad g=1,\ldots,G
\end{aligned}
\end{equation}

where \(D_{g}\) is the diffusion coefficient for group g, \(\phi_{g}\) is the neutron flux in group g, \(\Sigma_{\mathrm{r},g}\) is the removal cross section, \(\Sigma_{\mathrm{s},g' \rightarrow g}\) is the macroscopic scattering cross section from group g$'$ to group g, \({\nu\Sigma}_{\mathrm{f},g'}\) is the fission neutron production cross section, and \(\mathcal{X}_{g}\) is the fission spectrum.

In the above K-eigenvalue equation, both \(k_{eff}\) and the neutron flux are involved in the steady-state neutron balance of the system. When \(k_{eff}\) changes, the fission source term changes accordingly, and the corresponding neutron flux distribution satisfying the steady-state balance also changes. Therefore, \(k_{eff}\) and the neutron flux need to be determined simultaneously during the solution process.

To achieve the simultaneous solution of the neutron flux and \(k_{eff}\), JI-PINN uses a neural network to approximate the neutron flux in each energy group and treats \(k_{eff}\) as an independent trainable parameter. The neutron fluxes in all energy groups are represented as

\begin{equation}
\label{eq:2}
\begin{array}{r}
\widehat{\Phi}\left( x;\ \mathbf{\theta} \right) = \left\lbrack {\widehat{\phi}}_{1}\left( x;\ \mathbf{\theta} \right),{\widehat{\phi}}_{2}\left( x;\ \mathbf{\theta} \right),...,{\widehat{\phi}}_{G}\left( x;\ \mathbf{\theta} \right) \right\rbrack
\end{array}
\end{equation}

where \(\mathbf{\theta}\) denotes the flux network parameters.

To obtain initialization information at relatively low computational cost, JI-PINN constructs a discrete K-eigenvalue problem on a low-resolution spatial grid to obtain approximate solutions for the neutron flux and \(k_{eff}\):

\begin{equation}
\label{eq:3}
\begin{array}{r}
A_{h}{\widetilde{\Phi}}_{h} = \frac{1}{{\widetilde{k}}_{eff,h}}F_{h}{\widetilde{\Phi}}_{h}
\end{array}
\end{equation}

where h denotes the spatial discretization scale, \(A_{h}\) and \(F_{h}\) denote the discretized non-fission and fission operators, respectively, and \({\widetilde{\Phi}}_{h}\) and \({\widetilde{k}}_{eff,h}\) constitute the approximate K-eigenvalue solution of the discrete problem.

The discrete flux \({\widetilde{\Phi}}_{h}\) is spatially interpolated to obtain a continuous representation of the approximate flux, \(\widetilde{\Phi}(x)\). The initial values of the flux network parameters are then constructed by fitting \(\widetilde{\Phi}(x)\), and the initialization loss is defined as

\begin{equation}
\label{eq:4}
\begin{array}{r}
\mathcal{L}_{\mathrm{pre}} = \frac{1}{N_{\mathrm{p}}G}\sum_{i = 1}^{N_{\mathrm{p}}}{\sum_{g = 1}^{G}\left| {\widehat{\phi}}_{g}\left( x_{i};\mathbf{\theta} \right) - {\widetilde{\phi}}_{g}\left( x_{i} \right) \right|^{2}}
\end{array}
\end{equation}

By minimizing Eq.~\eqref{eq:4}, the initial values of the flux network parameters \(\mathbf{\theta}^{(0)}\) are obtained, and \(k_{eff}^{(0)} = {\widetilde{k}}_{eff,h}\) is simultaneously set. The two constitute the joint initial state of JI-PINN. Subsequently, \((\mathbf{\theta}^{(0)},k_{eff}^{(0)})\) is used as the starting point for training, and \(\mathbf{\theta}\) and \(k_{eff}\) are jointly optimized under physical constraints. With Eq.~\eqref{eq:2} substituted into Eq.~\eqref{eq:1}, the physics residual of the neutron diffusion equation for group g is defined as

\begin{equation}
\label{eq:5}
\begin{array}{r}
\mathcal{R}_{g}\left( x;\mathbf{\theta},k_{eff} \right) = \mathcal{N}_{g}\left\lbrack \widehat{\Phi}\left( x;\mathbf{\theta} \right),k_{eff} \right\rbrack,\quad g = 1,\ldots,G
\end{array}
\end{equation}

Based on the governing equation residual in Eq.~\eqref{eq:5}, together with the boundary conditions and the flux scaling constraint, the physics-constrained loss of JI-PINN is written as

\begin{equation}
\label{eq:6}
\begin{array}{r}
\mathcal{L}_{\mathrm{PINN}} = \lambda_{\mathrm{f}}\mathcal{L}_{\mathrm{f}} + \lambda_{\mathrm{b}}\mathcal{L}_{\mathrm{b}} + \lambda_{\mathrm{s}}\mathcal{L}_{\mathrm{s}}
\end{array}
\end{equation}

where \(\mathcal{L}_{\mathrm{f}}\), \(\mathcal{L}_{\mathrm{b}}\), and \(\mathcal{L}_{\mathrm{s}}\) denote the governing equation, boundary condition, and flux scaling losses, respectively. The flux network parameters \(\mathbf{\theta}\) and \(k_{eff}\) are jointly optimized by minimizing this loss, and the neutron flux distribution and \(k_{eff}\) are ultimately obtained.

\begin{equation}
\label{eq:7}
\begin{array}{r}
\left( \mathbf{\theta}^{*},k_{eff}^{\mathbf{*}} \right) = \underset{\mathbf{\theta},k_{eff}}{\operatorname*{arg\,min}}{\mathcal{L}_{\mathrm{PINN}}\left( \mathbf{\theta},k_{eff} \right)}
\end{array}
\end{equation}

\section{Experiments}
\label{sec:experiments}
In this section, JI-PINN was evaluated from four perspectives: the quality of initialization information, the role of joint initialization, sensitivity to random initialization, and computational performance under different problem conditions. These aspects were examined separately by considering the trade-off between initialization information quality and additional computational cost; the effects of the initial values of the flux network parameters and \(k_{eff}\) on subsequent optimization and their joint effect; solution stability under different random initial states; and performance in terms of efficiency and accuracy with increasing material heterogeneity and changes in spatial dimensionality.

At the level of test cases, this study selected the two-dimensional two-group two-material case to analyze the effects of the spatial resolution of the approximate solution and the degree of flux network pretraining on initialization effectiveness and computational cost, and evaluated the stability of the results across multiple random seeds \cite{ref19}. For the IAEA 2D benchmark, an initialization ablation study was conducted to distinguish the effects of the initial values of the flux network parameters and the initial \(k_{eff}\) \cite{ref15}. The two-dimensional two-group four-material case \cite{ref20} and the three-dimensional single-group cubic case were further used to evaluate the computational performance of the method under changes in material heterogeneity and spatial dimensionality \cite{ref19}.

\subsection{Two-Dimensional Two-Group Two-Material Case}

The computational domain and material distribution of the two-dimensional two-group two-material case are shown in Fig. 2, and the material parameters are listed in Table I. The computational domain consisted of two material regions, and zero flux boundary conditions were imposed on all outer boundaries. To fix the amplitude of the neutron flux, two fast group neutron flux scaling constraint points were placed at the material interface at \((0, - 0.5)\) and \((0,0.5)\).

\begin{figure}[tbp]
\centering
\includegraphics[width=0.82\columnwidth]{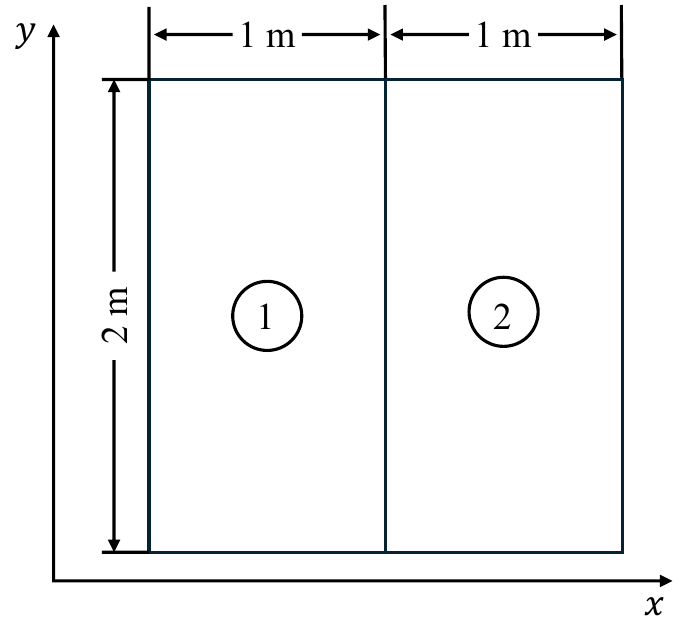}
\caption{Geometry and material distribution of the two-dimensional two-group two-material case.}
\label{fig:2}
\end{figure}

\begin{table}[tbp]
\caption{Material parameters of the two-dimensional two-group two-material case}
\label{tab:material-two}
\centering\scriptsize
\setlength{\tabcolsep}{1.2pt}
\begin{tabular*}{\columnwidth}{@{\extracolsep{\fill}}cccccc@{}}
\hline
Material & $g$ & $D_g\,(\mathrm{cm})$ & $\Sigma_{\mathrm{a}}\,(\mathrm{cm}^{-1})$ & $\nu\Sigma_{\mathrm{f}}\,(\mathrm{cm}^{-1})$ & $\Sigma_{1\rightarrow2}\,(\mathrm{cm}^{-1})$ \\
\hline
\multirow{2}{*}{Material 1} & 1 & 1.268 & 0.007181 & 0.004609 & 0.02767 \\
 & 2 & 0.1902 & 0.07047 & 0.08675 & $-$ \\
\multirow{2}{*}{Material 2} & 1 & 1.255 & 0.008252 & 0.004602 & 0.02533 \\
 & 2 & 0.211 & 0.1003 & 0.1091 & $-$ \\
\hline
\end{tabular*}
\end{table}

This case first investigated the relationship between the quality of initialization information and computational cost. The spatial resolution \(N_{h}\) of the approximate solution denotes the number of nodes of the low-resolution discretization grid along each corresponding spatial direction and determines the spatial information of the flux used for network initialization. The number of pretraining steps \(M_{\mathrm{pre}}\) controls the degree to which the network fits the approximate flux. A one-factor-at-a-time approach was adopted to separately investigate the effects of these two factors on initialization effectiveness. Based on preliminary tests for this case, \(N_{h}\) was set to 17, 25, 33, and 49 to cover a resolution range from insufficient spatial information to basic preservation of the main flux structure. At lower resolutions, the main spatial features of the flux were difficult to represent adequately. As the resolution was further increased, the improvement in the initialization state provided by additional local information gradually diminished, while also incurring higher computational costs for approximate solution calculation and flux network pretraining.

\(M_{\mathrm{pre}}\) was set to 1000, 3000, 5000, and 8000 to examine the process by which the network progressively formed the main spatial distribution of the approximate neutron flux as the degree of pretraining increased from an insufficient level. With fewer pretraining steps, the network fitted the approximate flux only to a limited extent and had difficulty forming a stable initial flux representation. As pretraining continued, the network gradually captured the main spatial features of the approximate neutron flux. However, longer pretraining increased the computational cost and could also lead to greater retention of the discretization and interpolation deviations in the low-resolution approximate solution.

The initialization effectiveness was evaluated using the initial loss \(L_{0}\) at the start of physics-constrained training, the total computational time \(T_{\mathrm{total}}\), and the final relative error of \(k_{eff}\). \(T_{\mathrm{total}}\) included the computational time required for approximate solution calculation, flux network pretraining, and subsequent physics-constrained training. The final relative error of \(k_{eff}\) was defined as

\begin{equation}
\label{eq:8}
\begin{array}{r}
\varepsilon_{\mathrm{k}} = \frac{\left| k_{eff}^{\mathrm{PINN}} - k_{eff}^{\mathrm{ref}} \right|}{\left| k_{eff}^{\mathrm{ref}} \right|}
\end{array}
\end{equation}

The different parameter settings and corresponding computational results are listed in Table II.

\begin{table}[tbp]
\caption{Results of the Hyperparameter Analysis for the Two-Dimensional Two-Group Two-Material Case}
\label{tab:hyperparameters}
\centering\scriptsize
\setlength{\tabcolsep}{1.2pt}
\begin{tabular*}{\columnwidth}{@{\extracolsep{\fill}}ccccc@{}}
\hline
Experimental Variable & Parameter Value & $L_{0}$ & $T_{\mathrm{total}}\,(\mathrm{s})$ & $\varepsilon_{\mathrm{k}}$ \\
\hline
\multirow{4}{*}{$N_{h}$} & 17 & $4.929\times10^{-3}$ & 797.9 & $6.693\times10^{-3}$ \\
 & 25 & $3.974\times10^{-3}$ & 1575.3 & $2.532\times10^{-3}$ \\
 & 33 & $5.946\times10^{-3}$ & 1631.0 & $2.730\times10^{-3}$ \\
 & 49 & $7.665\times10^{-3}$ & 1688.4 & $2.825\times10^{-3}$ \\
\multirow{4}{*}{$M_{\mathrm{pre}}$} & 1000 & $4.699\times10^{-1}$ & 8045.4 & $0.133\times10^{-3}$ \\
 & 3000 & $7.413\times10^{-3}$ & 1986.1 & $2.93\times10^{-3}$ \\
 & 5000 & $3.974\times10^{-3}$ & 1575.3 & $2.53\times10^{-3}$ \\
 & 8000 & $2.249\times10^{-3}$ & 831.7 & $5.41\times10^{-3}$ \\
\hline
\end{tabular*}
\par\vspace{2pt}
\parbox{\columnwidth}{\scriptsize\textit{Note:} The reference value is \(k_{eff}^{\mathrm{ref}} = 1.0695\).}
\end{table}

When \(N_{h}\) increased from 17 to 25, both \(L_{0}\) and \(\varepsilon_{\mathrm{k}}\) decreased markedly. The higher spatial resolution preserved the main spatial structure of the approximate flux more adequately and improved the initial function represented by the network at the start of physics-constrained training. When \(N_{h}\) was further increased to 33 and 49, neither \(L_{0}\) nor \(\varepsilon_{\mathrm{k}}\) decreased further, while \(T_{\mathrm{total}}\) continued to increase. At this stage, the main spatial features governing the flux distribution had already been largely preserved. Further increasing the resolution mainly added local spatial information, provided only limited improvement to the training starting point, and increased the computational overhead of approximate solution calculation and flux network pretraining. Therefore, the spatial resolution of the approximate solution needed to balance the completeness of the initial flux information against the additional computational overhead.

\(M_{\mathrm{pre}}\) controlled the degree of fit between the approximate flux and the initial function representation of the network. When \(M_{\mathrm{pre}}\)=1000, the network had not yet sufficiently formed the main spatial distribution of the approximate neutron flux, and substantial adjustment of the flux function was still required during subsequent physics-constrained training. As \(M_{\mathrm{pre}}\) increased, the initial function represented by the network gradually approached the spatial distribution of the approximate flux. When \(M_{\mathrm{pre}}\) was further increased from 5000 to 8000, \(L_{0}\) decreased further, whereas \(\varepsilon_{\mathrm{k}}\) increased. Longer pretraining allowed the network to fit the low-resolution approximate solution more fully, but could also retain the discretization and interpolation deviations contained in the approximate solution. These deviations still needed to be corrected during subsequent physics-constrained training.

Considering the initialization effectiveness, total computational time, final error, and training termination status, \(N_h\)=25 and \(M_{\mathrm{pre}}\)=5000 were used in the subsequent experiments. This combination retained effective initial flux information while avoiding the additional computational overhead caused by further increasing the spatial resolution and the potential loss of accuracy caused by prolonged pretraining. On this basis, 15 random seeds were selected for the comparison of the randomly initialized PINN, $R^{2}$-PINN \cite{ref13}, and JI-PINN. The statistical results are listed in Table III, and the distributions of the results for the three methods are shown in Fig. 3.

\begin{table*}[t]
\caption{Statistical Results of the Random Seed Experiments for the Two-Dimensional Two-Group Two-Material Case}
\label{tab:random-seeds}
\centering\scriptsize
\setlength{\tabcolsep}{3.2pt}
\begin{tabular*}{\textwidth}{@{\extracolsep{\fill}}cccccc@{}}
\hline
Method & Final $k_{eff}$ & $\varepsilon_{\mathrm{k}}$ & Fast Group MSE & Thermal Group MSE & $T_{\mathrm{total}}\,(\mathrm{s})$ \\
\hline
Randomly Initialized PINN & $1.037\pm0.039$ & $(3.141\pm3.512)\times10^{-2}$ & $(1.088\pm1.299)\times10^{-1}$ & $(1.634\pm1.835)\times10^{-2}$ & $3309.6\pm366.2$ \\
$R^{2}$-PINN & $1.056\pm0.027$ & $(1.252\pm2.532)\times10^{-2}$ & $(2.276\pm5.791)\times10^{-2}$ & $(3.389\pm8.454)\times10^{-3}$ & $4033.4\pm806.2$ \\
JI-PINN & $1.066\pm0.001$ & $(2.428\pm1.511)\times10^{-3}$ & $(5.615\pm3.246)\times10^{-3}$ & $(8.330\pm4.892)\times10^{-4}$ & $2467.7\pm931.5$ \\
\hline
\end{tabular*}
\par\vspace{2pt}
\parbox{\textwidth}{\scriptsize\textit{Note:} The reference value is \(k_{eff}^{\mathrm{ref}} = 1.0695\). The data in the table are reported as the mean \(\pm\) standard deviation over 15 random seed runs.}
\end{table*}

\begin{figure*}[t]
\centering
\includegraphics[width=0.80\textwidth]{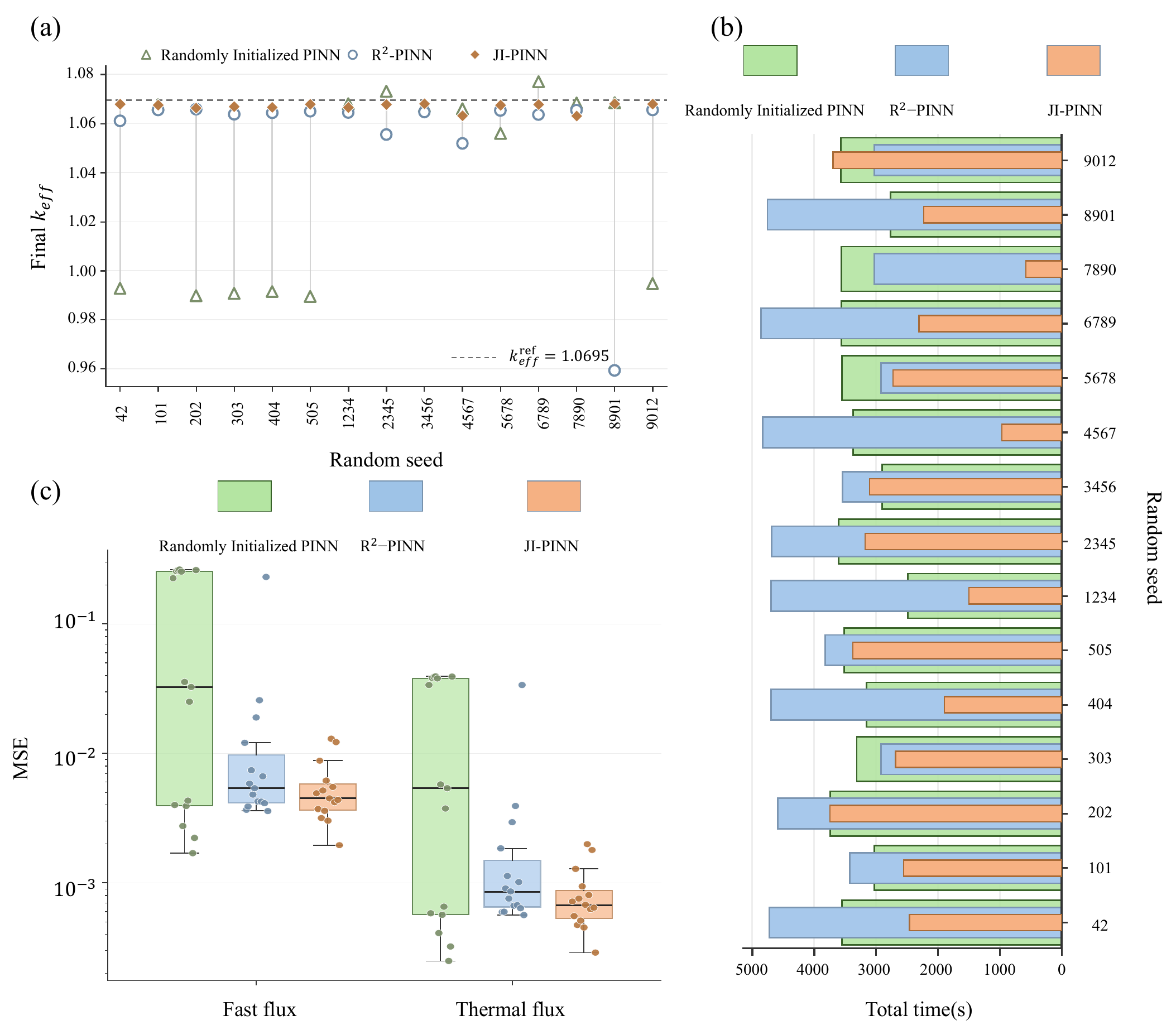}
\caption{Solution results and computational time for different random seeds: (a) final \(k_{eff}\); (b) total computational time; (c) MSE of the neutron flux for the fast and thermal groups.}
\label{fig:3}
\end{figure*}

Table III and Fig. 3 showed that the \(k_{eff}\) values, flux errors, and total computational times obtained with the randomly initialized PINN exhibited substantial variability, and the \(k_{eff}\) values obtained under some random seeds deviated markedly from the reference value. The $R^{2}$-PINN results showed a narrower distribution, but a few relatively large deviations still remained. The \(k_{eff}\) values obtained by JI-PINN across all 15 random seeds were concentrated near the reference value, and the variability in the flux errors of the fast and thermal groups was further reduced.

In terms of computational efficiency, the average total computational time of JI-PINN was reduced by 25.4\% and 38.8\% relative to the randomly initialized PINN and $R^{2}$-PINN, respectively, and remained at a relatively low level. As shown in Fig. 3(c), the mean squared error (MSE) of JI-PINN for the fast and thermal groups remained low overall, and the reduction in computational time was not accompanied by a noticeable loss of flux accuracy.

Different random seeds corresponded to different network parameter states and further produced substantially different initial flux functions. Although the flux networks in JI-PINN also started pretraining with randomly initialized parameters, all networks were subject to the combined effects of the same approximate flux distribution, boundary conditions, and flux scaling constraint. As a result, the differences among the flux functions at the start of formal physics-constrained training were reduced, while \(k_{eff}\) was initialized using the value obtained together with the approximate flux. Joint initialization made the initial states of networks with different random initializations more similar at the start of formal training, and the corresponding \(k_{eff}\), flux errors, and computational times also exhibited more concentrated distributions across different random seeds.

Based on the hyperparameter analysis, the quality of the approximate flux information and the degree of pretraining jointly determined the initial state provided by joint initialization. An appropriate spatial resolution and degree of pretraining could preserve the main spatial features of the fluxTNS while controlling the additional computational overhead, thereby balancing initialization quality and overall computational efficiency.

\FloatBarrier
\subsection{IAEA 2D Benchmark}

A quarter-core model was used for the IAEA 2D benchmark, with a computational domain of 170 cm \(\times\) 170 cm containing four material regions. Neumann boundary conditions were imposed on the left and bottom boundaries, while vacuum boundary conditions were imposed on the remaining stepped outer boundaries. The geometry and boundary types are shown in Fig. 4, and the material parameters are listed in Table IV.

\begin{figure}[!h]
\centering
\includegraphics[width=0.95\columnwidth]{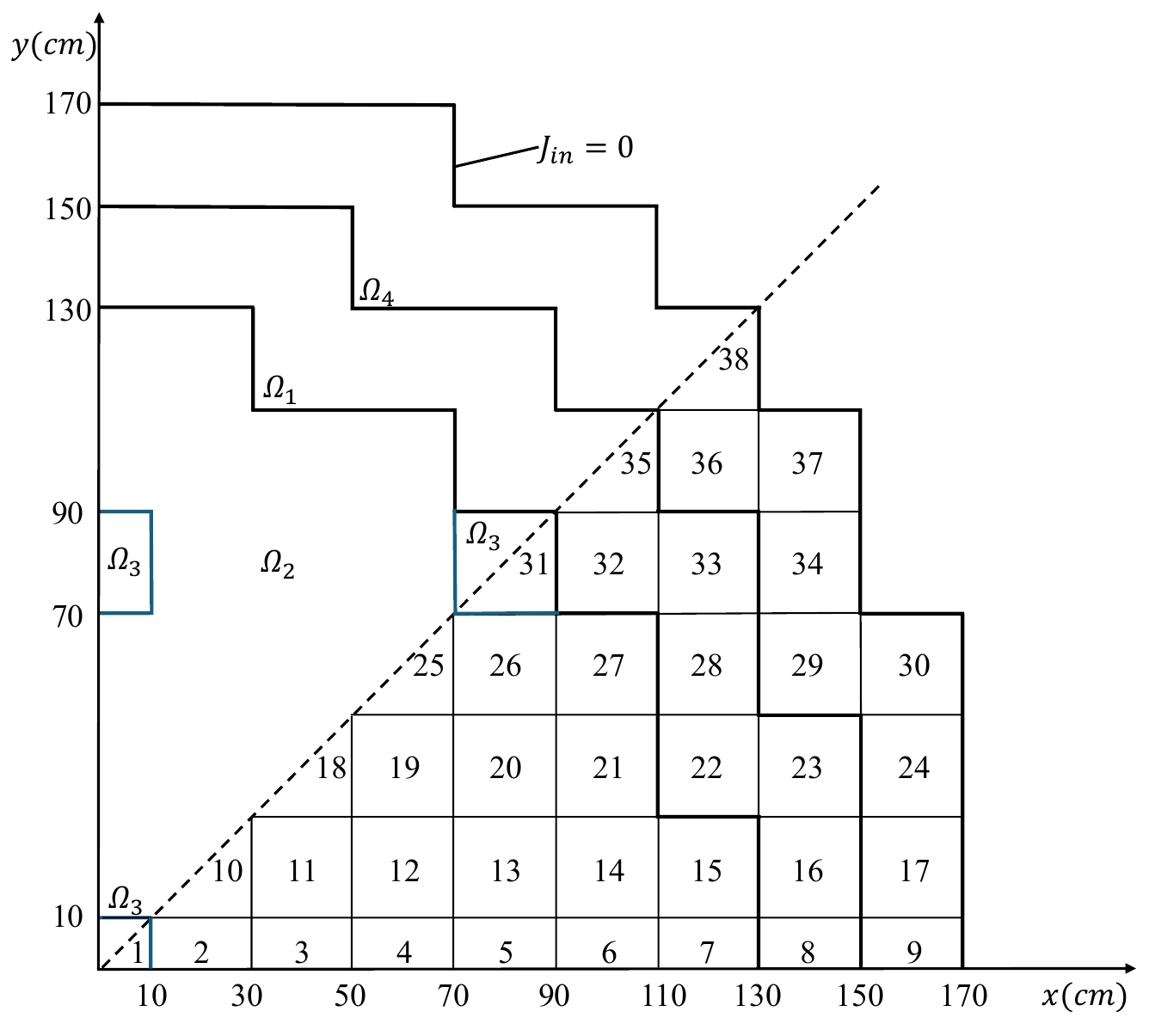}
\begingroup
\CLASSOPTIONconferencetrue
\caption{Geometric layout of the IAEA 2D benchmark.}
\label{fig:4}
\endgroup
\end{figure}

\begin{table}[!h]
\caption{Material parameters of the IAEA 2D Benchmark}
\label{tab:iaea-material}
\centering\scriptsize
\setlength{\tabcolsep}{1.2pt}
\begin{tabular*}{\columnwidth}{@{\extracolsep{\fill}}cccccc@{}}
\hline
Region & $g$ & $D_g\,(\mathrm{cm})$ & $\Sigma_{\mathrm{a}}\,(\mathrm{cm}^{-1})$ & $\nu\Sigma_{\mathrm{f}}\,(\mathrm{cm}^{-1})$ & $\Sigma_{1\rightarrow2}\,(\mathrm{cm}^{-1})$ \\
\hline
\multirow{2}{*}{$\Omega_{1}$} & 1 & 1.5 & 0.010 & 0.000 & 0.02 \\
 & 2 & 0.4 & 0.080 & 0.135 & $-$ \\
\multirow{2}{*}{$\Omega_{2}$} & 1 & 1.5 & 0.010 & 0.000 & 0.02 \\
 & 2 & 0.4 & 0.085 & 0.135 & $-$ \\
\multirow{2}{*}{$\Omega_{3}$} & 1 & 1.5 & 0.010 & 0.000 & 0.02 \\
 & 2 & 0.4 & 0.130 & 0.135 & $-$ \\
\multirow{2}{*}{$\Omega_{4}$} & 1 & 2.0 & 0.000 & 0.000 & 0.04 \\
 & 2 & 0.3 & 0.010 & 0.000 & $-$ \\
\hline
\end{tabular*}
\end{table}

The neutron fluxes in the fast and thermal groups were approximated by a neural network with two outputs. An ablation study on initialization was conducted in this case to examine the roles of the initial flux representation of the network and the initial value of \(k_{eff}\) in subsequent joint optimization. The flux network was either randomly initialized or pretrained using the approximate flux, while \(k_{eff}^{(0)}\) was set to either 1.0 or the approximate K-eigenvalue \({\widetilde{k}}_{eff}\). The
computational results are listed in Table V.

\begin{center}
\begin{minipage}{\columnwidth}
\centering
\captionof{table}{Computational Results for the IAEA 2D Benchmark under Different Initialization Configurations}
\label{tab:iaea-ablation}
\scriptsize
\setlength{\tabcolsep}{0.6pt}
\begin{tabular*}{\columnwidth}{@{\extracolsep{\fill}}cccccc@{}}
\hline
Initialization Method & $k_{eff}^{(0)}$ & $k_{eff}$ & $L_{0}$ & $\varepsilon_{\mathrm{k}}$ & $T_{\mathrm{total}}\,(\mathrm{s})$ \\
\hline
\multirow{2}{*}{Random Initialization} & 1 & 1.029586 & $6.420\times10^{5}$ & $3.48\times10^{-4}$ & 7398.5 \\
 & 1.029240 & 1.030006 & $6.420\times10^{5}$ & $6.02\times10^{-5}$ & 7106.3 \\
\multirow{2}{*}{Flux Initialization} & 1 & 1.030104 & $8.858\times10^{3}$ & $1.55\times10^{-4}$ & 5194.9 \\
 & 1.029240 & 1.029367 & $8.842\times10^{3}$ & $5.61\times10^{-4}$ & 4573.0 \\
\hline
\end{tabular*}
\par\vspace{2pt}
\parbox{\columnwidth}{\scriptsize\textit{Note:} The reference value is \(k_{eff}^{\mathrm{ref}} = 1.029944\).}
\end{minipage}
\end{center}

The final \(k_{eff}\) values obtained under all four initialization configurations remained close to the reference value, but clear differences were observed in the training starting points and total computational times. Under random flux initialization, when \(k_{eff}^{(0)}\) was adjusted from 1.0 to \({\widetilde{k}}_{eff,h}\), \(T_{\mathrm{total}}\) decreased only from 7398.5 s to 7106.3 s, a reduction of 3.9\%. In this case, the flux network still entered physics-constrained training from a random function state, and the spatial distribution of the flux still required gradual adjustment during subsequent optimization. Therefore, improving only the initial value of \(k_{eff}\) provided relatively limited time savings.

After approximate flux pretraining, with the same setting of \(k_{eff}^{(0)}\)=1, \(L_{0}\) decreased markedly, and \(T_{\mathrm{total}}\) decreased from 7398.5 s to 5194.9 s, a reduction of 29.8\%. Before formal physics-constrained optimization began, the pretrained network already exhibited the main spatial distribution of the approximate flux, reducing the adjustment required for the spatial structure of the flux function during subsequent training. On this basis, \({\widetilde{k}}_{eff,h}\), obtained together with the approximate flux, was further used as the initial value of \(k_{eff}\). The initial loss remained nearly unchanged, while \(T_{\mathrm{total}}\) decreased from 5194.9 s to 4573.0 s, a further reduction of 12.0\%.

Fig. 5 further showed the convergence histories of \(k_{eff}\) under the four initialization configurations. Random flux initialization was used in both Fig. 5(a) and Fig. 5(b). In Fig. 5(b), \(k_{eff}\) started from a value closer to the reference value, but noticeable iterative adjustment was still observed during the early stage of training. 

\begin{figure*}[t]
\centering
\includegraphics[width=0.84\textwidth]{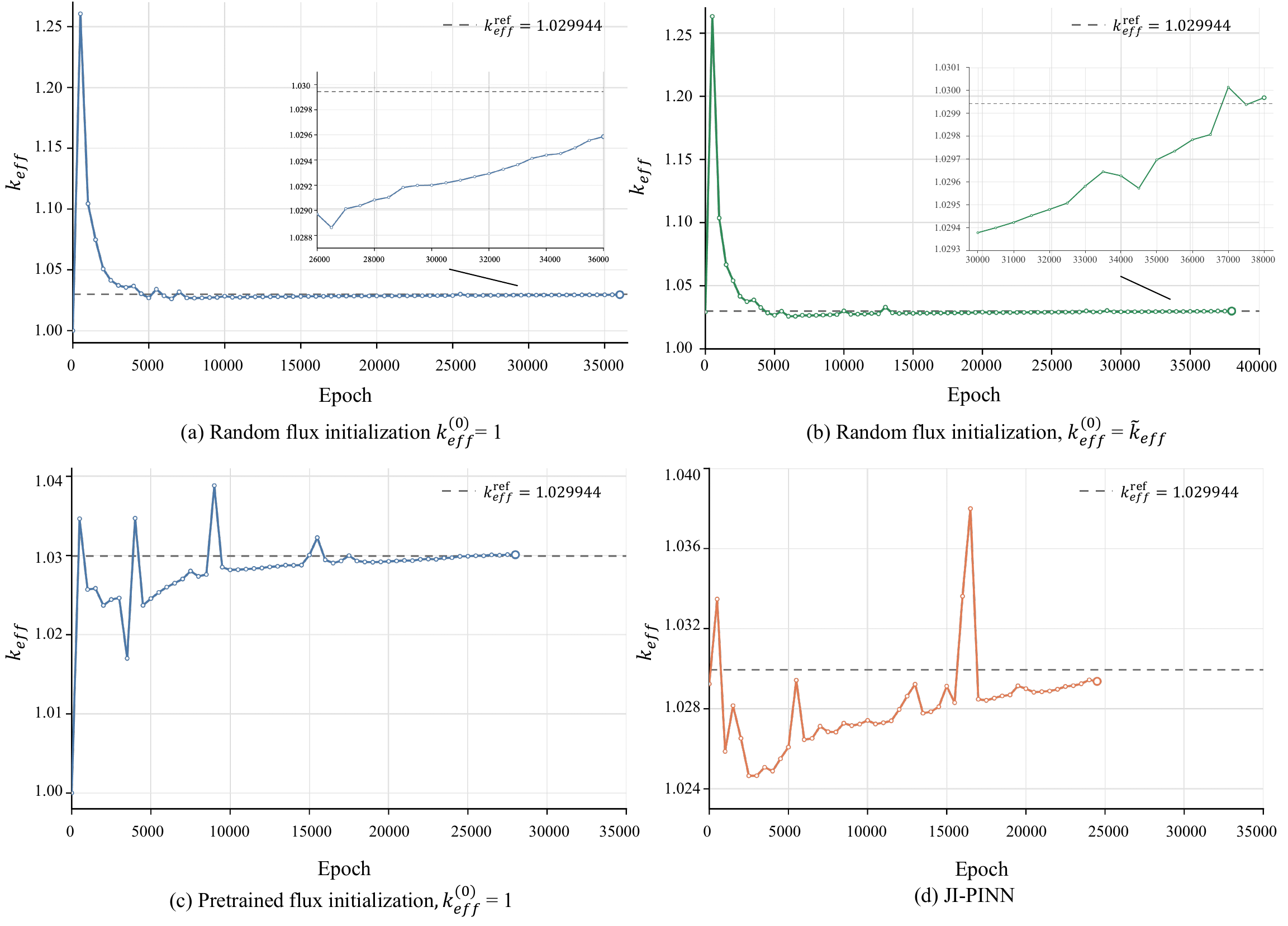}
\begingroup
\CLASSOPTIONconferencetrue
\caption{Convergence histories of \(k_{eff}\) under different initialization configurations.}
\label{fig:5}
\endgroup
\end{figure*}

Approximate flux pretraining was used in both Fig. 5(c) and Fig. 5(d). The corresponding \(k_{eff}\) convergence histories entered the vicinity of the reference value earlier than those under random flux initialization. Under the same pretrained flux condition, JI-PINN further used the corresponding \(k_{eff}\) obtained from that K-eigenvalue solution as the initial value of \(k_{eff}\), and \(T_{\mathrm{total}}\) was further reduced.

Considering the results from the four initialization ablation cases, approximate flux pretraining improved the flux function representation at the start of formal physics-constrained training. After the \(k_{eff}\) value obtained together with the approximate flux from the same approximate K-eigenvalue solution was further used as the initial value of \(k_{eff}\), the initial flux representation and the initial value of \(k_{eff}\) were both derived from the same approximate solution, forming an initial state more favorable for subsequent joint optimization.

\subsection{Two-Dimensional Two-Group Four-Material Case}

This case was used to investigate the computational performance of the proposed JI-PINN in a heterogeneous multi-material domain. The geometry and material distribution of the two-dimensional two-group four-material case are shown in Fig. 6, and the material parameters are listed in Table VI. Zero flux boundary conditions were imposed on the outer boundaries.

\begin{figure}[tbp]
\centering
\includegraphics[width=0.95\columnwidth]{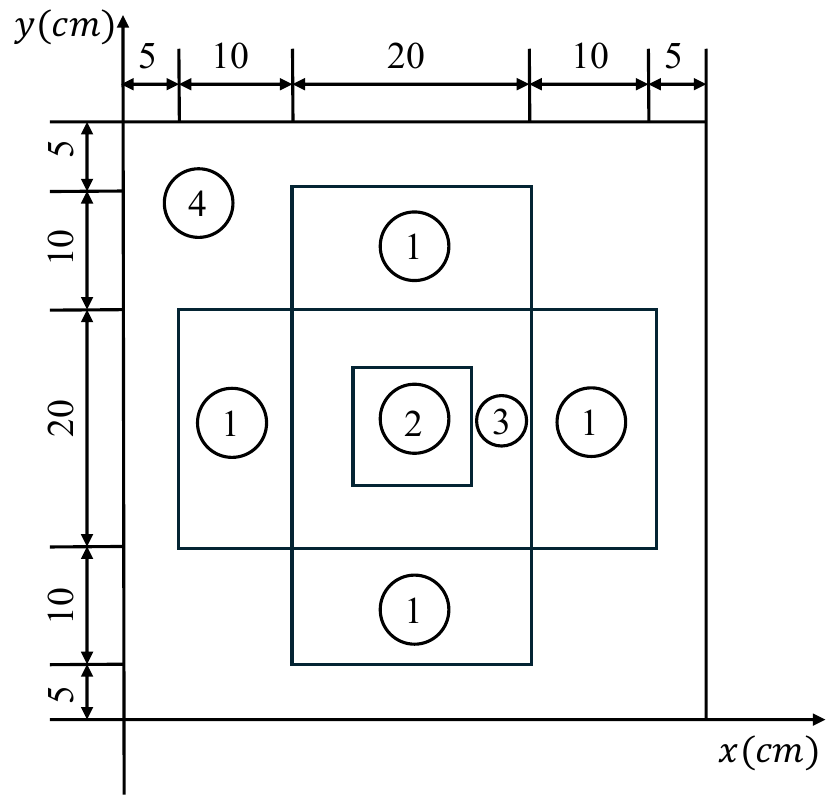}
\caption{Geometry and material distribution of the two-dimensional two-group four-material case.}
\label{fig:6}
\end{figure}

\begin{table}[!t]
\caption{Material parameters of the two-dimensional two-group four-material case}
\label{tab:material-four}
\centering\scriptsize
\setlength{\tabcolsep}{1.2pt}
\begin{tabular*}{\columnwidth}{@{\extracolsep{\fill}}cccccc@{}}
\hline
Material & $g$ & $D_g\,(\mathrm{cm})$ & $\Sigma_{\mathrm{a}}\,(\mathrm{cm}^{-1})$ & $\nu\Sigma_{\mathrm{f}}\,(\mathrm{cm}^{-1})$ & $\Sigma_{1\rightarrow2}\,(\mathrm{cm}^{-1})$ \\
\hline
\multirow{2}{*}{Material 1} & 1 & 1.255 & 0.008252 & 0.004602 & 0.02533 \\
 & 2 & 0.2110 & 0.100300 & 0.10910 & $-$ \\
\multirow{2}{*}{Material 2} & 1 & 1.268 & 0.007181 & 0.004609 & 0.02767 \\
 & 2 & 0.1902 & 0.070470 & 0.08675 & $-$ \\
\multirow{2}{*}{Material 3} & 1 & 1.259 & 0.008002 & 0.004663 & 0.02617 \\
 & 2 & 0.2091 & 0.083440 & 0.10210 & $-$ \\
\multirow{2}{*}{Material 4} & 1 & 1.259 & 0.008002 & 0.004663 & 0.02617 \\
 & 2 & 0.2091 & 0.073324 & 0.10210 & $-$ \\
\hline
\end{tabular*}
\end{table}

The neutron fluxes in the fast and thermal groups were approximated separately by independent fully connected neural networks. The computational results are listed in Table VII, and the convergence history of \(k_{eff}\) is shown in Fig. 7.

\begin{table}[!b]
\caption{Computational Results for the Two-Dimensional Two-Group Four-Material Case}
\label{tab:four-material-results}
\centering\scriptsize
\setlength{\tabcolsep}{1.2pt}
\begin{tabular*}{\columnwidth}{@{\extracolsep{\fill}}cccc@{}}
\hline
Method & $k_{eff}$ & $\varepsilon_{\mathrm{k}}$ & $T_{\mathrm{total}}\,(\mathrm{s})$ \\
\hline
Randomly Initialized PINN & 0.796108 & $4.88\times10^{-4}$ & 2757.6 \\
$R^{2}$-PINN & 0.795609 & $1.39\times10^{-4}$ & 4578.0 \\
JI-PINN & 0.793601 & $2.66\times10^{-3}$ & 1394.9 \\
\hline
\end{tabular*}
\par\vspace{2pt}
\parbox{\columnwidth}{\scriptsize\textit{Note:} The reference value is \(k_{eff}^{\mathrm{ref}} = 0.79572\).}
\end{table}

\begin{figure}[tbp]
\centering
\includegraphics[width=\columnwidth,trim=12bp 2bp 4bp 2bp,clip]{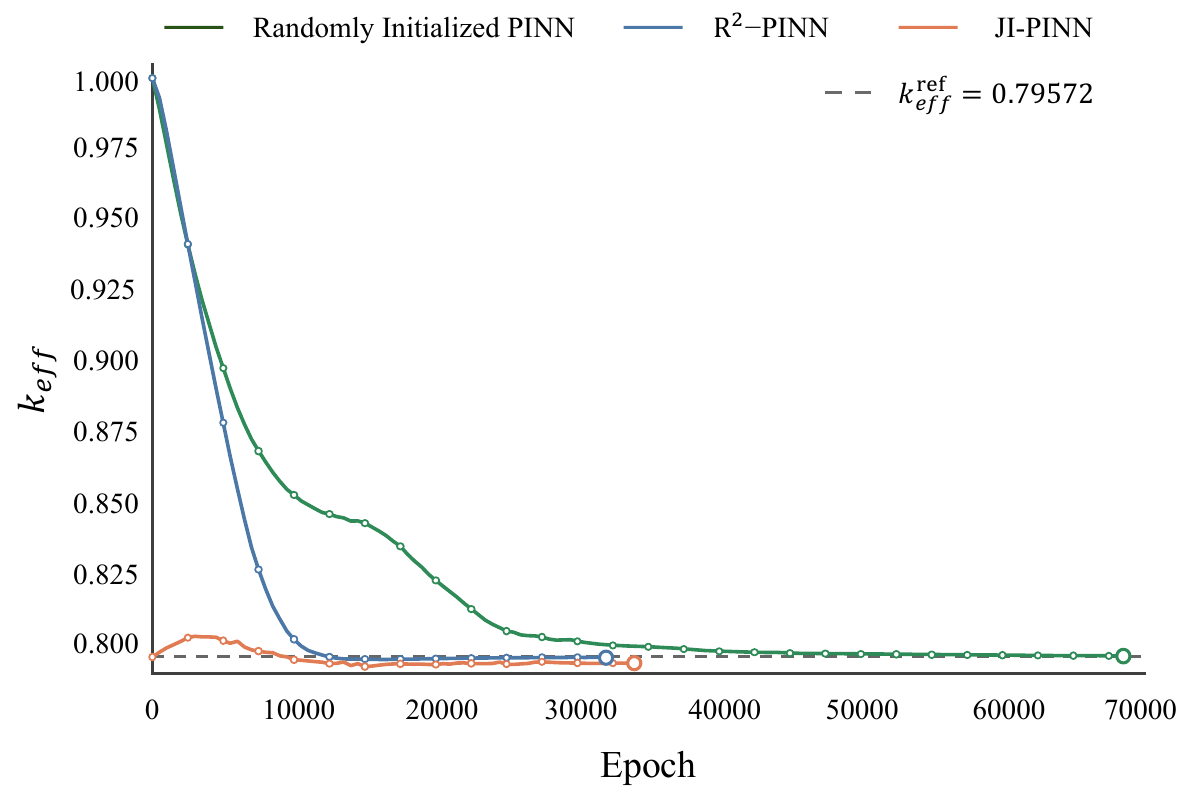}
\caption{Convergence history of \(k_{eff}\) for the two-dimensional two-group four-material case.}
\label{fig:7}
\end{figure}

As shown in Table VII, \(T_{\mathrm{total}}\) of JI-PINN was reduced by 49.4\% and 69.5\% compared with those of the randomly initialized PINN and $R^{2}$-PINN, respectively. The convergence histories in Fig. 7 showed that, for the randomly initialized PINN and $R^{2}$-PINN, \(k_{eff}\) had to be progressively adjusted from an empirically set initial value toward the reference value. In contrast, JI-PINN entered physics-constrained training with an initial state closer to the reference value, exhibited only small fluctuations during the early stage of training, and then entered a stable range more rapidly. The improved initial state reduced the extent of adjustment required for the flux representation and \(k_{eff}\) during the early stage of training, thereby shortening \(T_{\mathrm{total}}\). However, the relative error of \(k_{eff}\) for JI-PINN was higher than those of the two comparison methods. The shorter subsequent training was insufficient to fully correct the local deviations retained in the approximate initial state, and the improvement in computational efficiency was accompanied by some loss of accuracy.

As the number of material interfaces increased, local flux variations near the interfaces became more pronounced, and the low-resolution approximate solution was limited in its ability to represent these local features, so some local deviations could still remain after joint initialization. The extent to which these deviations were corrected during subsequent physics-constrained training further affected the final solution accuracy.

\FloatBarrier
\begin{table}[!t]
\caption{Computational Results for the Three-Dimensional Single-Group Cubic Case}
\label{tab:cubic-results}
\centering\scriptsize
\setlength{\tabcolsep}{1.0pt}
\begin{tabular*}{\columnwidth}{@{\extracolsep{\fill}}ccccc@{}}
\hline
Method & $k_{eff}$ & $\varepsilon_{\mathrm{k}}$ & Flux MSE & $T_{\mathrm{total}}\,(\mathrm{s})$ \\
\hline
Randomly Initialized PINN & 0.9505335 & $5.62\times10^{-4}$ & $6.15\times10^{-7}$ & 697.4 \\
$R^{2}$-PINN & 0.950510 & $5.37\times10^{-4}$ & $8.17\times10^{-7}$ & 612.6 \\
JI-PINN & 0.950660 & $6.95\times10^{-4}$ & $9.49\times10^{-7}$ & 496.1 \\
\hline
\end{tabular*}
\par\vspace{2pt}
\parbox{\columnwidth}{\scriptsize\textit{Note:} The reference value is \(k_{eff}^{\mathrm{ref}} = 0.9500\).}
\end{table}

\subsection{Three-Dimensional Single-Group Cubic Case}

The computational domain of this case was a cube with a side length of 1 m, and zero flux boundary conditions were imposed on all six outer surfaces, with \(k_{\infty}\)=1.425 and the square of the diffusion length \(L^{2}\)=0.0168869. A flux scaling constraint of \(\phi(0,0,0) = 0.5\) was also imposed at the center of the cube, and the geometry is shown in Fig. 8.

\begin{figure}[!htbp]
\centering
\includegraphics[width=0.78\columnwidth]{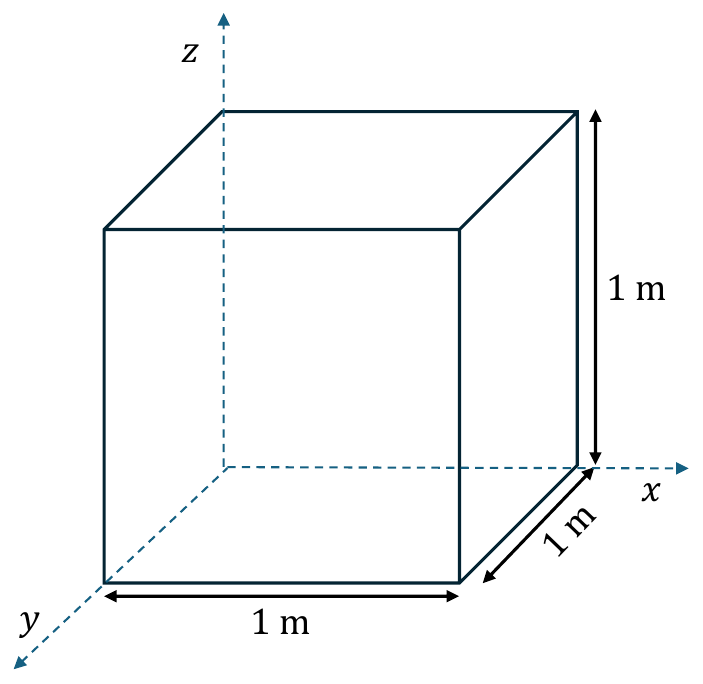}
\begingroup
\CLASSOPTIONconferencetrue
\caption{Geometry of the three-dimensional single-group cubic case.}
\label{fig:8}
\endgroup
\end{figure}

The computational results are listed in Table VIII, and the convergence history of \(k_{eff}\) is shown in Fig. 9.

\begin{figure*}[t]
\centering
\includegraphics[width=0.98\textwidth]{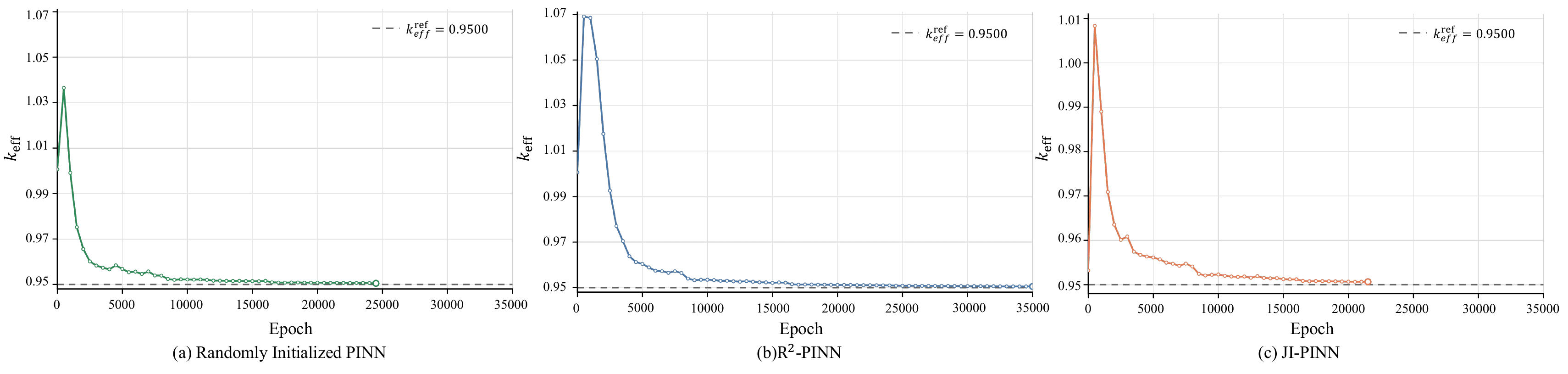}
\begingroup
\CLASSOPTIONconferencetrue
\caption{Convergence history of \(k_{eff}\) for the three-dimensional single-group cubic case.}
\label{fig:9}
\endgroup
\end{figure*}

In Table VIII, the relative errors of \(k_{eff}\) and the neutron flux MSE obtained by the three methods were of the same order of magnitude, and the final solution accuracies were generally comparable. The \(T_{\mathrm{total}}\) of JI-PINN was approximately 28.9\% and 19.0\% lower than those of the randomly initialized PINN and $R^{2}$-PINN, respectively. In Fig. 9, both the randomly initialized PINN and $R^{2}$-PINN underwent relatively large adjustments of \(k_{eff}\) during the early stage of training. JI-PINN used approximate flux pretraining to set the initial values of the network, so that the flux representation and the trainable parameter \(k_{eff}\) were optimized starting from the existing approximate flux representation and its corresponding initial \(k_{eff}\) value, thereby reducing the correction range during the early stage of training and satisfying the stopping criterion with fewer training steps. However, brief fluctuations were still observed during the early iterations, indicating that the low-resolution approximate state still required further correction under physical constraints.

In the three-dimensional setting, joint initialization still improved the initial state for formal training and reduced the overall computational time while maintaining comparable solution accuracy.

\section{Conclusion}
\label{sec:conclusion}
For the joint solution of the neutron flux and \(k_{eff}\) in steady-state neutron diffusion K-eigenvalue problem, this paper proposes JI-PINN, a method based on joint initialization of the flux network and \(k_{eff}\), and validates it through extensive numerical tests on the two-dimensional two-group two-material case, the IAEA 2D benchmark, the two-dimensional two-group four-material case, and the three-dimensional single-group case. The results from these cases showed that JI-PINN reduced the total computational time by 25.4\%--49.4\% compared with the randomly initialized PINN. The hyperparameter analysis and random seed experiments revealed the influence of the starting point of training on subsequent optimization. An appropriate spatial resolution of the approximate solution and degree of pretraining balanced the preservation of the main flux information against the additional computational cost and reduced the differences in the initial states of networks with different random initializations when entering formal training. The ablation study on initialization for the IAEA 2D benchmark further distinguished the roles of the two types of initialization information: the approximate flux improved the flux function representation at the start of formal training, while the corresponding \(k_{eff}\) provided an initial value closer to the target solution for eigenvalue optimization. Together, they formed an initial state more favorable for subsequent physics-constrained joint optimization. As the problems were extended to heterogeneous domains composed of multiple material regions and to three-dimensional settings, JI-PINN still exhibited a clear advantage in computational efficiency while maintaining a favorable balance between efficiency and accuracy.

Overall, JI-PINN provides a new approach to improving the overall solution efficiency of steady-state neutron diffusion K-eigenvalue problem. However, the computational performance of JI-PINN is still affected by the quality of the low-resolution approximate solution, especially for problems with complex material interfaces and pronounced local flux variations, where the ability of the approximate solution to represent local spatial information can further affect physics-constrained optimization and the final solution accuracy. Future work will further investigate the applicability of the method to more complex material regions, three-dimensional multigroup reactor core models, and repeated calculations for parameterized problems, and will explore its combination with other efficient physics-constrained solution strategies.

\IEEEtriggercmd{\enlargethispage{3\baselineskip}}
\IEEEtriggeratref{7}


\begin{thebibliography}{20}
\bibitem{ref1} J. J. Duderstadt and L. J. Hamilton, \emph{Nuclear Reactor Analysis}. New York, NY, USA: Wiley, 1976.

\bibitem{ref2} E. E. Lewis and W. F. Miller, Jr., \emph{Computational Methods of Neutron Transport}. New York, NY, USA: Wiley, 1984.

\bibitem{ref3} W. M. Stacey, \emph{Nuclear Reactor Physics}, 3rd ed. Weinheim, Germany: Wiley-VCH, 2018.

\bibitem{ref4} R. D. Lawrence, ``Progress in nodal methods for the solution of the neutron diffusion and transport equations,'' \emph{Prog. Nucl. Energy}, vol. 17, no. 3, pp. 271--301, 1986, doi: 10.1016/0149-1970(86)90034-X.

\bibitem{ref5} A. Hébert, \emph{Applied Reactor Physics}. Montréal, QC, Canada: Presses Internationales Polytechnique, 2009.

\bibitem{ref6} P. Cao, C. Cao, and Q. Gan, ``A 3-D neutron distribution reconstruction method based on the off-situ measurement for reactor,'' \emph{IEEE Trans. Nucl. Sci.}, vol. 68, no. 12, pp. 2694--2701, Dec. 2021, doi: 10.1109/TNS.2021.3123381.

\bibitem{ref7}
X. Chen and A. Ray, ``Deep reinforcement learning control of a boiling water reactor,''
\emph{IEEE Trans. Nucl. Sci.},
vol. 69, no. 8, pp. 1820--1832, Aug. 2022, doi: 10.1109/TNS.2022.3187662.

\bibitem{ref8} I. E. Lagaris, A. Likas, and D. I. Fotiadis, ``Artificial neural networks for solving ordinary and partial differential equations,'' \emph{IEEE Trans. Neural Netw.}, vol. 9, no. 5, pp. 987--1000, Sep. 1998, doi: 10.1109/72.712178.

\bibitem{ref9} M. Raissi, P. Perdikaris, and G. E. Karniadakis, ``Physics-informed neural networks: A deep learning framework for solving forward and inverse problems involving nonlinear partial differential equations,'' \emph{J. Comput. Phys.}, vol. 378, pp. 686--707, Feb. 2019, doi: 10.1016/j.jcp.2018.10.045.

\bibitem{ref10} H. Zhang, J. Li, Y. Yi, Q. Hang, and Y. Jiang, ``Gradient harmony optimization-driven hybrid physics-informed neural network for solving neutron diffusion problems,'' \emph{IEEE Trans. Nucl. Sci.}, vol. 73, no. 3, pp. 505--518, Mar. 2026, doi: 10.1109/TNS.2026.3655477.

\bibitem{ref11} J. Wang, X. Peng, Z. Chen, B. Zhou, Y. Zhou, and N. Zhou, ``Surrogate modeling for neutron diffusion problems based on conservative physics-informed neural networks with boundary conditions enforcement,'' \emph{Ann. Nucl. Energy}, vol. 176, Oct. 2022, Art. no. 109234, doi: 10.1016/j.anucene.2022.109234.

\bibitem{ref12} M.-H. Do, K. Ammar, N. G. Castaing, and F. Madiot, ``Physics informed neural networks for the mixed dual form of the neutron diffusion equation with heterogeneous coefficients,'' \emph{Ann. Nucl. Energy}, vol. 223, Dec. 2025, Art. no. 111607, doi: 10.1016/j.anucene.2025.111607.

\bibitem{ref13} H. Zhang, Y.-L. He, D. Liu, Q. Hang, H.-M. Yao, and D. Xiang, ``Residual resampling-based physics-informed neural network for neutron diffusion equations,'' \emph{Nucl. Sci. Tech.}, vol. 37, no. 2, Feb. 2026, Art. no. 28, doi: 10.1007/s41365-025-01839-5.

\bibitem{ref14} M. H. Elhareef and Z. Wu, ``Physics-informed neural network method and application to nuclear reactor calculations: A pilot study,'' \emph{Nucl. Sci. Eng.}, vol. 197, no. 4, pp. 601--622, Apr. 2023, doi: 10.1080/00295639.2022.2123211.

\bibitem{ref15} Y. Yang \emph{et al}., ``A data-enabled physics-informed neural network with comprehensive numerical study on solving neutron diffusion eigenvalue problems,'' \emph{Ann. Nucl. Energy}, vol. 183, Apr. 2023, Art. no. 109656, doi: 10.1016/j.anucene.2022.109656.

\bibitem{ref16} Q.-H. Yang, Y. Yang, Y.-T. Deng, Q.-L. He, H.-L. Gong, and S.-Q. Zhang, ``Physics-constrained neural network for solving discontinuous interface K-eigenvalue problem with application to reactor physics,'' \emph{Nucl. Sci. Tech.}, vol. 34, no. 10, Oct. 2023, Art. no. 161, doi: 10.1007/s41365-023-01313-0.

\bibitem{ref17} H. Bi, M. Song, T. Zhang, and X. Liu, ``FC-PINNs: Physics-informed neural networks for solving neutron diffusion eigenvalue problem with interface considerations,'' \emph{J. Comput. Phys.}, vol. 541, Nov. 2025, Art. no. 114311, doi: 10.1016/j.jcp.2025.114311.

\bibitem{ref18} J. C. Wong, C. C. Ooi, A. Gupta, and Y.-S. Ong, ``Learning in sinusoidal spaces with physics-informed neural networks,'' \emph{IEEE Trans. Artif. Intell.}, vol. 5, no. 3, pp. 985--1000, Mar. 2024, doi: 10.1109/TAI.2022.3192362.

\bibitem{ref19} D. Liu, L. Tang, P. An, B. Zhang, and Y. Jiang, ``The deep learning method to search effective multiplication factor of nuclear reactor directly,'' (in Chinese), \emph{Nucl. Power Eng.}, vol. 44, no. 5, pp. 6--14, Oct. 2023, doi: 10.13832/j.jnpe.2023.05.0006.

\bibitem{ref20} Y. Jiang, P. An, D. Liu, and Y. Yu, ``Research on the solution and acceleration algorithm of source iteration method based on PINN,'' (in Chinese), \emph{Nucl. Power Eng.}, vol. 46, no. 2, pp. 148--155, Apr. 2025, doi: 10.13832/j.jnpe.2024.090040.
\end{thebibliography}
\end{document}